%% file: main_scirep_v1.tex
\documentclass[10pt]{article}

\usepackage[margin=0.75in]{geometry}
\usepackage{amsmath,amssymb}
\usepackage{graphicx}
\usepackage{booktabs}
\usepackage{multirow}
\usepackage{hyperref}
\hypersetup{hidelinks}
\usepackage{cite}
\usepackage{enumitem}
\usepackage{float}
\usepackage{url}
\usepackage{setspace}
\usepackage[table]{xcolor}

\title{Standalone LLM and a Pre-specified Agentic Pipeline for Explaining ICU Mortality Predictions: a Feasibility Study on the eICU Demo Dataset}

\author{
Di Zhu\textsuperscript{1,a}
\and
Chen Xie\textsuperscript{1,b}
\and
Haoyun Zhang\textsuperscript{c}
\and
Zihan Wei\textsuperscript{d}
\and
Ziwei Wang\textsuperscript{*,d}
\and
Jiazhao Shi\textsuperscript{e}
\and
Ziyu Wang\textsuperscript{f}
\and
Qiyang Xie\textsuperscript{g}
}

\date{
\small
\textsuperscript{a}Santa Clara University, Santa Clara, United States\\
\textsuperscript{b}University of Massachusetts Amherst, Amherst, United States\\
\textsuperscript{c}University of Pennsylvania, Philadelphia, United States\\
\textsuperscript{d}Carnegie Mellon University, Pittsburgh, United States\\
\textsuperscript{e}New York University, Brooklyn, United States\\
\textsuperscript{f}Wake Forest University, Winston-Salem, United States\\
\textsuperscript{g}Northeastern University, Boston, United States\\
\textsuperscript{1}These authors contributed equally.\\
\textsuperscript{*}Corresponding author: Ziwei Wang.
}

\begin{document}

\maketitle
\setstretch{1.0}

\begin{abstract}
Machine-learning models can predict ICU mortality accurately, but feature-attribution methods alone rarely provide the clinical narrative needed for bedside use. Large language models (LLMs) may bridge this gap, and multi-step agentic pipelines are a plausible extension because they separate data interpretation, guideline checking, and final explanation. This revised feasibility study preserves the original standalone-versus-agentic comparison while making the main clinical findings more explicit. Using the retained local eICU Demo artifact set (2,353 ICU stays; 8.1\% mortality), XGBoost achieved an AUROC of 0.855 (95\% CI 0.796--0.906) and an AUPRC of 0.332 (95\% CI 0.217--0.494). On a stratified 38-case explanation subset, the standalone LLM produced 1 explanation with explicit outcome leakage, whereas the four-step agentic pipeline produced none. Among the 14 cases that overlapped with the SHAP review subset, the standalone LLM showed higher SHAP alignment (mean Jaccard 0.171 versus 0.077) and higher direction consistency (92.9\% versus 78.6\%), while the agentic pipeline showed higher guideline grounding (0.762 versus 0.143), higher value specificity (0.236 versus 0.143), and slightly higher plausibility (0.700 versus 0.671). Clinically, the results suggest that agentic decomposition may improve safety-relevant grounding and patient-specific detail, but it should be paired with attribution-based checks before use in high-stakes risk explanation.
\end{abstract}

\section{Introduction}

Mortality prediction in the intensive care unit (ICU) remains a central benchmark for critical-care risk modeling. Modern machine-learning models often exceed the discrimination of classical severity scores when applied to structured electronic health record data \cite{Johnson2017,harutyunyan2019}. The practical barrier is not purely predictive performance; it is whether clinicians can understand and audit why a model has assigned high risk to a particular patient \cite{tonekaboni2019}.

Post-hoc explainability methods such as SHAP are valuable because they identify the features that move a model prediction \cite{lundberg2017,lundberg2020}. However, feature attribution is not equivalent to clinical explanation. A ranked list of variables, even when mathematically faithful, does not by itself produce the pathophysiological narrative that clinicians use for verification, triage, and communication \cite{ghassemi2021}. This gap has motivated interest in large language models (LLMs), which can transform structured values into concise natural-language reasoning grounded in clinical concepts \cite{singhal2023,nori2023,lee2023}.

A single LLM prompt is an attractive baseline, but it may collapse multiple reasoning steps into one opaque response. Agentic systems offer an alternative by decomposing the task into data interpretation, application of formal criteria, differential construction, and synthesis \cite{wang2024agents,xi2024agents}. That decomposition is especially appealing in critical care, where thresholds, syndromic criteria, and multiple interacting organ systems matter. Our original study design therefore asked three research questions:

\begin{enumerate}[leftmargin=2em]
    \item[\textbf{RQ1}] How accurately can standard machine-learning models predict ICU mortality from structured first-24-hour data?
    \item[\textbf{RQ2}] Can a standalone LLM produce clinically plausible explanations that retain some alignment with SHAP attributions?
    \item[\textbf{RQ3}] Does a structured agentic pipeline produce higher-quality, more clinically grounded explanations than a standalone LLM?
\end{enumerate}

This revised manuscript preserves the same questions and underlying idea, but it reports only claims supported by the auditable local artifact set. Specifically, RQ1 is addressed quantitatively on the held-out cohort, whereas RQ2 and RQ3 are addressed on an audited explanation subset generated from the same versioned test snapshot.

\section{Results}

\subsection{Cohort characteristics and predictive performance}

The study cohort contained 2,353 adult ICU stays with an in-hospital mortality rate of 8.1\% ($n=191$). Non-survivors were older and showed higher heart rate, respiratory rate, lactate, and blood urea nitrogen values than survivors. Baseline descriptive statistics are shown in Table~\ref{tab:demographics}.

\input{results/scirep_v1/tables/table1_demographics_scirep_v1.tex}

\begin{figure}[!htbp]
\centering
\includegraphics[width=0.72\textwidth]{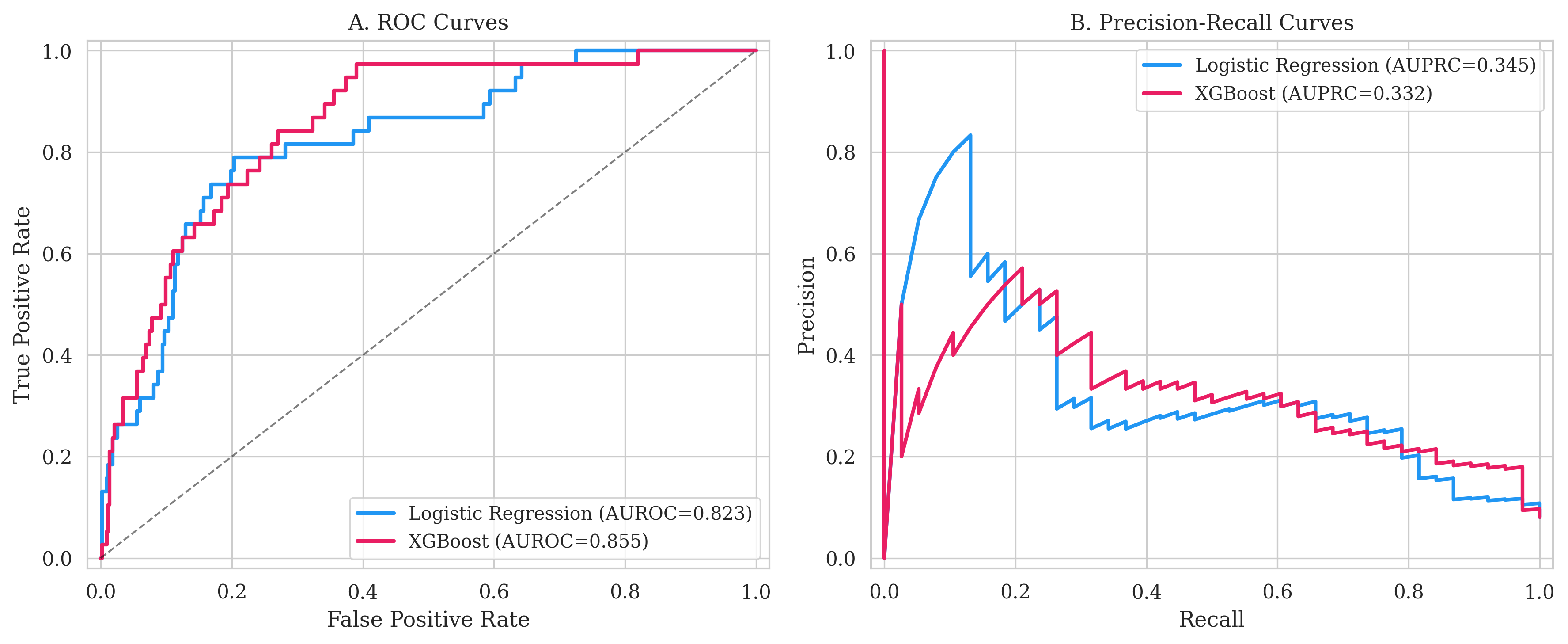}
\caption{Receiver operating characteristic and precision-recall curves for logistic regression and XGBoost on the held-out test set.}
\label{fig:rocpr}
\end{figure}

XGBoost outperformed logistic regression on the held-out test set, although both models showed the typical precision-recall constraints expected under low event prevalence. XGBoost achieved an AUROC of 0.855 (95\% CI 0.796--0.906) and an AUPRC of 0.332 (95\% CI 0.217--0.494), whereas logistic regression achieved an AUROC of 0.823 (95\% CI 0.752--0.886) and an AUPRC of 0.345 (95\% CI 0.218--0.506) (Table~\ref{tab:model_performance}; Figure~\ref{fig:rocpr}). These discrimination estimates are consistent with prior ICU mortality modeling studies based on structured EHR data \cite{Johnson2017,harutyunyan2019}.

\input{results/scirep_v1/tables/table2_model_performance_scirep_v1.tex}

\subsection{Global feature attribution with SHAP}

\begin{figure}[!htbp]
\centering
\includegraphics[width=0.72\textwidth]{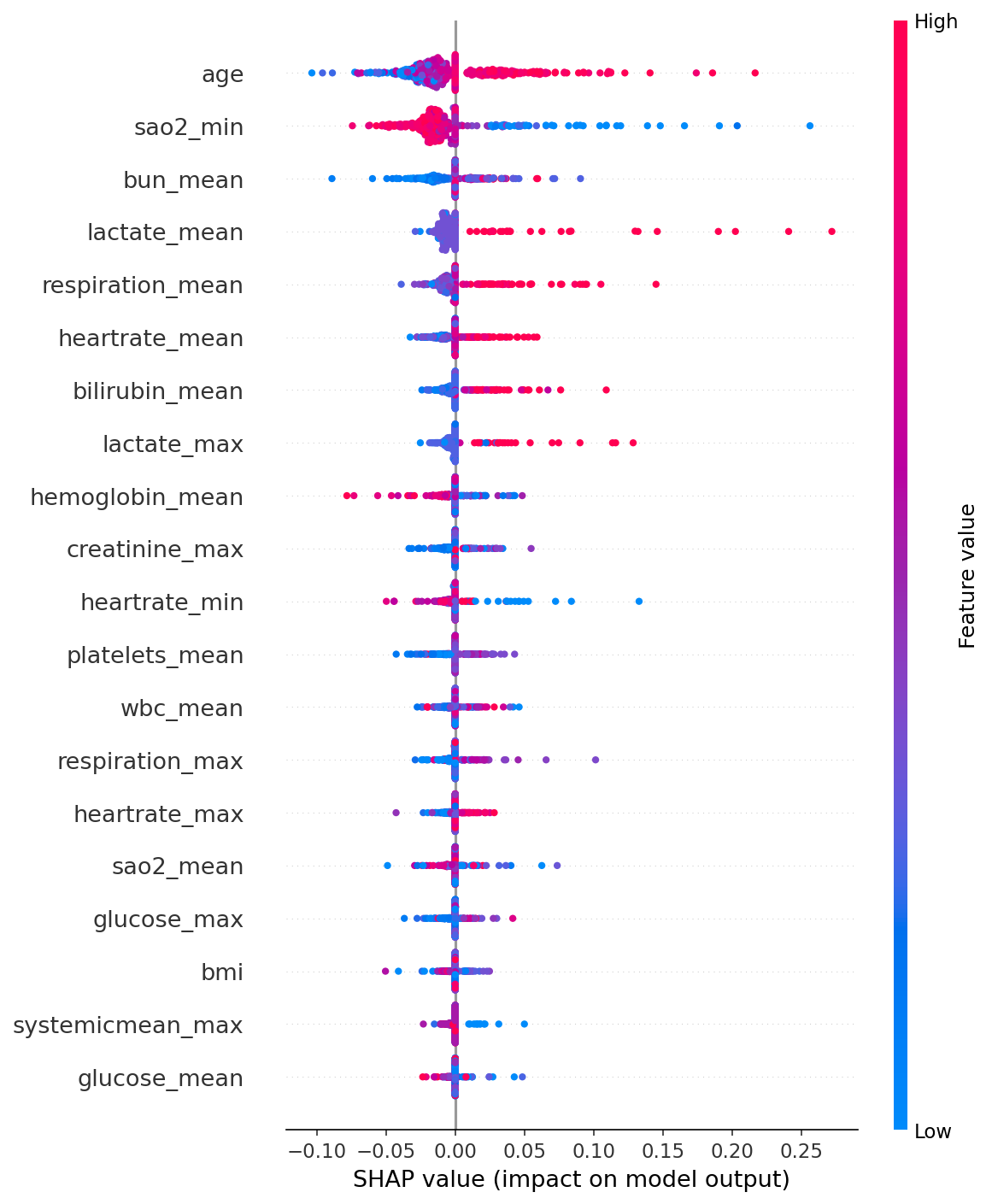}
\caption{SHAP summary plot for the XGBoost model. Each point represents one patient, and color denotes feature value.}
\label{fig:shap}
\end{figure}

The SHAP summary remained clinically coherent. Age, minimum SpO$_2$, blood urea nitrogen, lactate, and respiratory rate were the most influential features in the retained XGBoost model, matching well-established ICU mortality correlates such as advanced age, hypoxemia, renal dysfunction, and respiratory compromise \cite{gutierrez2020}. The summary plot is shown in Figure~\ref{fig:shap}.

\subsection{Standalone LLM audit and retained explanation quality}

The regenerated standalone explanation set contained 38 outputs. A direct audit of explanation text found 1 explanation with explicit outcome leakage terms and removed it from the valid comparison set. After this audit, 37 explanations remained, and 14 overlapped with the pre-existing SHAP-reviewed patient subset and could therefore be evaluated against feature attribution.

\begin{figure}[!htbp]
\centering
\includegraphics[width=0.72\textwidth]{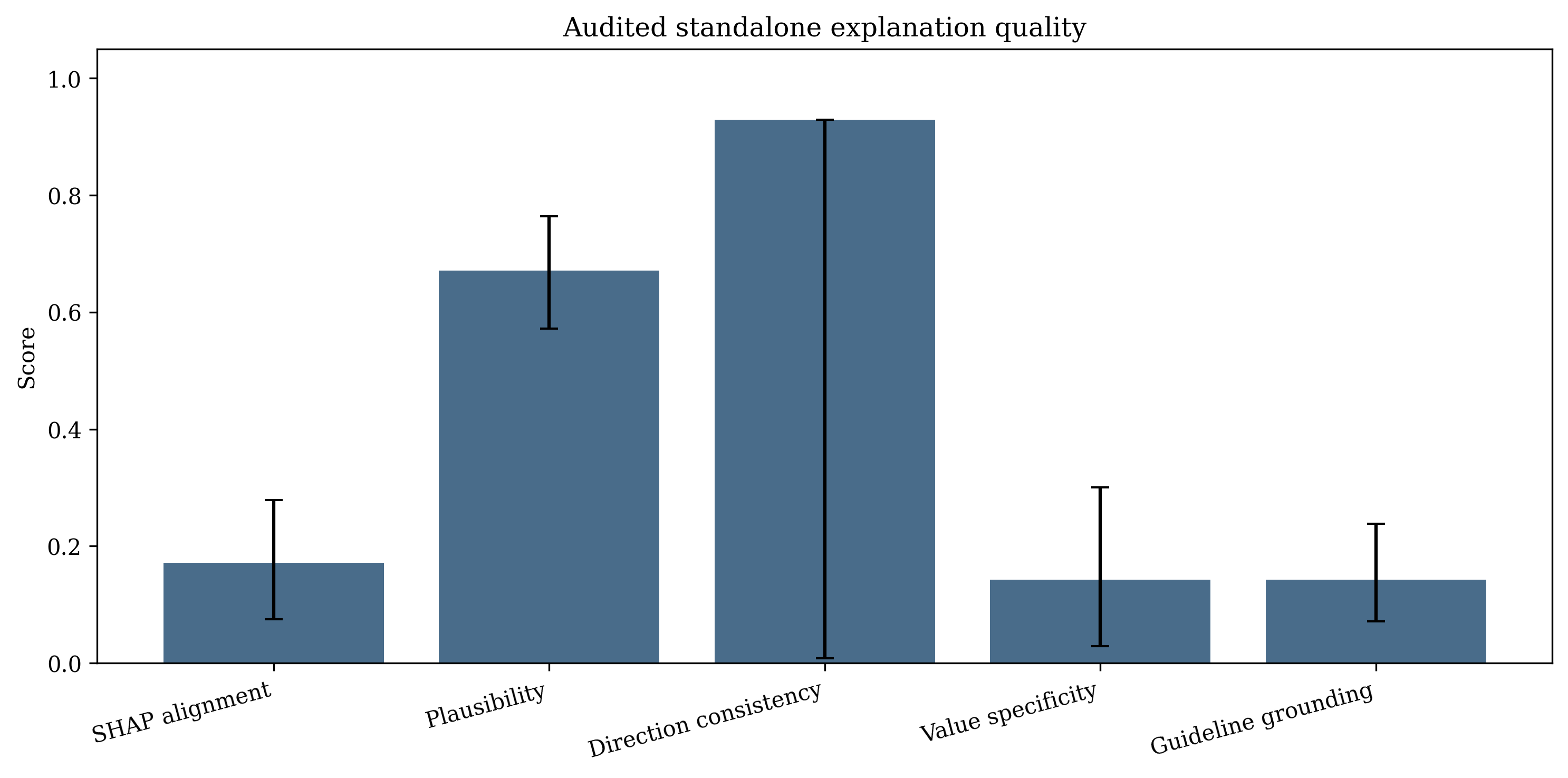}
\caption{Audited standalone explanation quality metrics with bootstrap confidence intervals. Direction consistency is normalized to the 0--1 scale for display.}
\label{fig:llmquality}
\end{figure}

Within these 14 leakage-free overlapping cases, explanation quality remained mixed (Table~\ref{tab:llm_quality}; Figure~\ref{fig:llmquality}). Mean SHAP alignment was 0.171 (95\% CI 0.075--0.279), and 50.0\% of cases showed any feature overlap at all. Explanation plausibility was moderate (mean 0.671, 95\% CI 0.571--0.764), while direction consistency with the model-predicted risk label remained high at 92.9\% (95\% CI 78.6--100.0). Value specificity was low (mean 0.143, 95\% CI 0.029--0.300), and guideline grounding was limited (mean 0.143, 95\% CI 0.071--0.238). Taken together, these results suggest that the standalone LLM can produce concise and directionally coherent narratives, but those narratives remain only modestly aligned with the model's top SHAP features.

\input{results/scirep_v1/tables/table3_explanation_quality_scirep_v1.tex}

\subsection{Agentic pipeline comparison}

The same four-step agentic design from the original manuscript was preserved in this refit because it is central to RQ3. The pipeline comprises a clinical data interpreter, a guideline consultant, a differential reasoner, and a final synthesizer. Each step was rewritten in the versioned scripts to avoid outcome leakage and to consume cleaned prompt-facing values only. The architecture is shown in Figure~\ref{fig:pipeline}.

\begin{figure}[!htbp]
\centering
\includegraphics[width=0.72\textwidth]{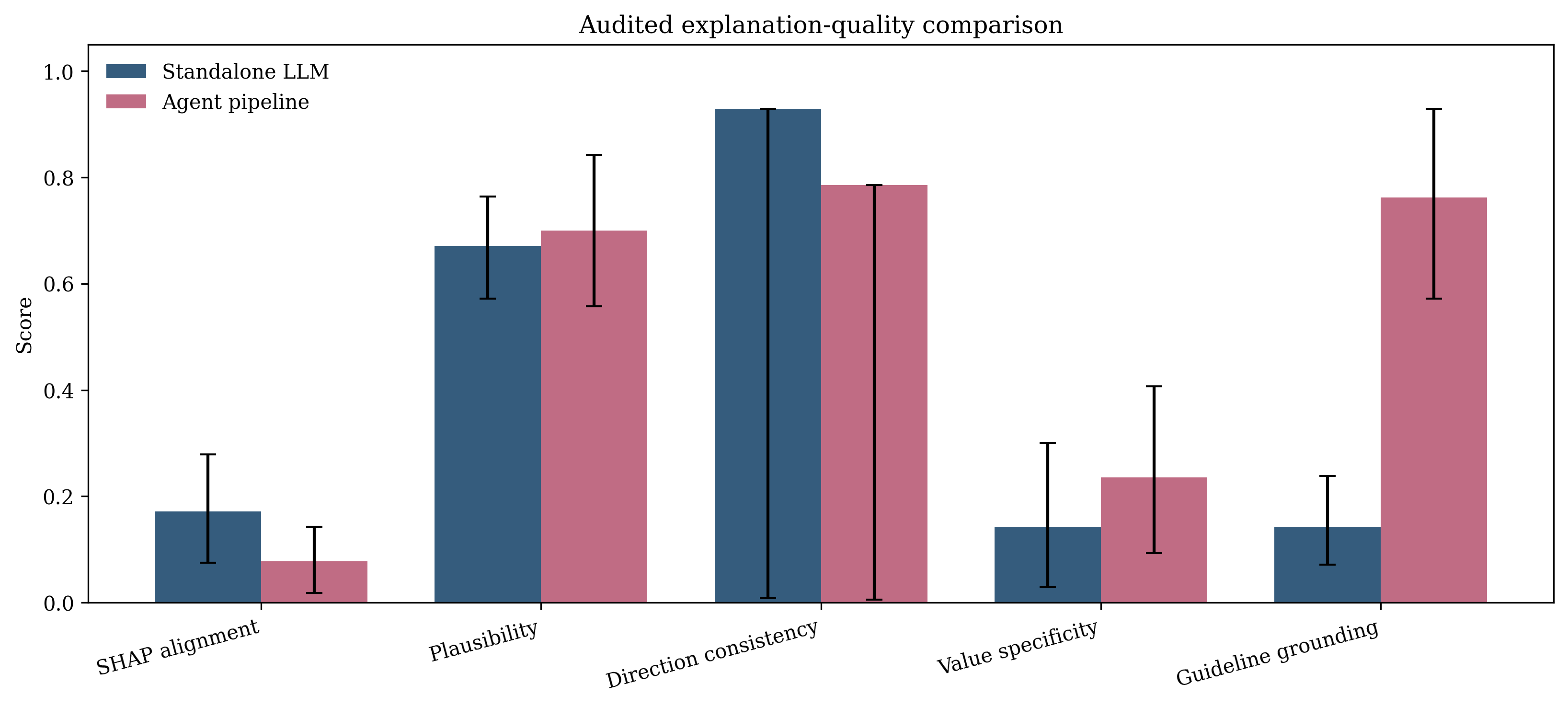}
\caption{Audited explanation-quality comparison between the standalone LLM and the pre-specified agentic pipeline. Direction consistency is normalized to the 0--1 scale for display.}
\label{fig:comparison}
\end{figure}

The regenerated agent run also produced 38 outputs, and none were flagged for explicit outcome leakage. Fourteen cases overlapped with the SHAP-reviewed subset and were therefore available for direct comparison with the regenerated standalone baseline (Table~\ref{tab:explanation_comparison}; Figure~\ref{fig:comparison}). The comparative pattern was not monotonic. The agentic pipeline had lower SHAP alignment than the standalone baseline (mean Jaccard 0.077, 95\% CI 0.018--0.143, versus 0.171, 95\% CI 0.075--0.279) and lower direction consistency (78.6\% versus 92.9\%). In contrast, it showed markedly stronger guideline grounding (0.762, 95\% CI 0.571--0.929, versus 0.143, 95\% CI 0.071--0.238), higher value specificity (0.236, 95\% CI 0.093--0.407, versus 0.143, 95\% CI 0.029--0.300), more frequent mention of patient-specific data (85.7\% versus 64.3\%), and slightly higher plausibility (0.700, 95\% CI 0.557--0.843, versus 0.671, 95\% CI 0.571--0.764). These results suggest that task decomposition made the explanations more explicit about clinical criteria and supporting evidence, but did not move them closer to the dominant SHAP features of the underlying mortality model.

\input{results/scirep_v1/tables/table4_explanation_comparison_scirep_v1.tex}

\begin{figure}[!htbp]
\centering
\scriptsize
\begin{tabular}{|p{0.92\textwidth}|}
\hline
\textbf{Step 1: Clinical data interpreter} \\
\textit{Input:} cleaned patient values plus reference ranges \\
\textit{Output:} abnormality summary with patient-specific values and clinical significance \\
\hline
$\downarrow$ \\
\hline
\textbf{Step 2: Guideline consultant} \\
\textit{Input:} patient data plus Step 1 summary \\
\textit{Output:} structured application of SIRS, SOFA, qSOFA, and KDIGO-style criteria \\
\hline
$\downarrow$ \\
\hline
\textbf{Step 3: Differential reasoner} \\
\textit{Input:} patient data plus Steps 1--2 outputs \\
\textit{Output:} ranked mortality-driving conditions with evidence and mechanisms \\
\hline
$\downarrow$ \\
\hline
\textbf{Step 4: Final synthesizer} \\
\textit{Input:} patient data plus Steps 1--3 outputs \\
\textit{Output:} structured JSON explanation in the same schema as the standalone baseline \\
\hline
\end{tabular}
\caption{Pre-specified outcome-free agentic pipeline retained for RQ3.}
\label{fig:pipeline}
\end{figure}

\section{Discussion}

The revised analysis prioritizes three clinically meaningful findings. First, the structured mortality model was sufficiently discriminative to justify explanation work, and its most influential features---age, oxygenation, blood urea nitrogen, lactate, and respiratory rate---are recognizable markers of physiologic instability in critical care. Second, the standalone LLM usually produced a concise risk narrative in the correct direction, but the explanations were often weak in patient-specific values and only modestly aligned with SHAP attributions. Third, the pre-specified agentic pipeline reduced explicit leakage and produced more guideline-grounded, patient-specific explanations, but this improvement came with lower SHAP alignment and lower direction consistency.

These findings support the original motivation while clarifying the practical tradeoff. Feature attribution and natural-language explanation can be complementary: SHAP helps identify what drove the model, whereas an LLM can translate abnormal values into a readable clinical narrative. However, an agentic pipeline should not be assumed to be superior simply because it is more structured. In this study, decomposition improved guideline use and bedside readability, but it did not automatically recover the features that most strongly drove the XGBoost prediction. A clinically useful system may therefore need both components: an attribution check for model fidelity and a guideline-grounded language layer for interpretability.

The revised analysis also highlights data-quality issues that directly affect explanation reliability. In the prompt-facing test snapshot, mean arterial pressure was missing in 84.7\% of rows after invalid values below 20 mmHg were excluded, temperature was missing in 93.4\% of rows after one Fahrenheit-style outlier was harmonized to Celsius, lactate was missing in 81.5\% of rows, and reconstructed GCS was unavailable in 13.2\% of rows. These values are clinically important, so missingness should be surfaced rather than hidden; otherwise, an LLM may sound more certain than the available evidence supports.

Several limitations remain. The eICU Demo dataset is small relative to full-scale ICU cohorts. Only 14 generated cases overlapped with the retained SHAP-reviewed subset, so the confidence intervals around the comparative explanation metrics remain wide. Both generators were run with a single local base model, so the observed differences reflect one particular implementation of standalone prompting and one particular implementation of task decomposition rather than an architecture-independent truth. The attribution comparison also depends on heuristic mapping from narrative factors to structured model features. Finally, automated explanation metrics remain proxies for clinician judgment, not substitutes for prospective evaluation by critical-care experts. Recent work on retrieval-grounded evaluation for conversational LLM-based risk assessment similarly emphasizes that risk-oriented LLM outputs should be judged by evidence grounding and task-relevant reference information, rather than fluency or plausibility alone \cite{hu2026retrieval}.

Despite those limitations, the revised manuscript strengthens the original conference draft by making the evidence more transparent. It adds an auditable head-to-head comparison, makes leakage explicit, separates prompt-cleaning rules from downstream interpretation, and presents agentic decomposition as a measurable tradeoff rather than a categorical gain.

\section{Methods}

\subsection{Data source and cohort}

We used the eICU Collaborative Research Database Demo v2.0.1 \cite{pollard2018}. The retained cohort definition from the original study included adults aged at least 18 years, ICU length of stay of at least 4 hours, and non-missing hospital discharge status. The final cohort contained 2,353 ICU stays, and the primary outcome was in-hospital mortality.

\subsection{Structured features and prompt-facing cleaning}

Features were derived from the first 24 hours and covered demographics, vital signs, laboratory results, and APACHE-related neurological variables. The versioned refit preserved the original model inputs but added prompt-facing cleaning rules for the explanation pipeline. GCS was reconstructed only from valid APACHE eye, motor, and verbal components; negative sentinels were not treated as clinical values. Temperature values above 45 were treated as Fahrenheit and converted to Celsius. Mean arterial pressure values below 20 mmHg and respiratory-rate values below 5 per minute were excluded from prompt-facing summaries. Missingness after these cleaning steps was tabulated in the versioned results tree.

\subsection{Mortality modeling}

The original predictive models were retained: L2-regularized logistic regression and XGBoost \cite{chen2016}. The scirep\_v1 refit added bootstrap 95\% confidence intervals for AUROC and AUPRC on the held-out test set by resampling the 471 test encounters with replacement.

\subsection{SHAP attribution}

The retained SHAP artifact set from the original study was reused to summarize global feature importance and to provide a per-patient explanation reference subset. SHAP top features for each reviewed patient were defined as the three largest absolute attributions.

\subsection{Standalone and agentic explanation designs}

The standalone baseline was defined as a single outcome-free prompt that received cleaned first-24-hour patient data and the model-predicted mortality probability, and returned a structured JSON explanation. The agentic pipeline used the same cleaned patient representation but decomposed the task into four serial steps: data interpretation, explicit guideline application, differential reasoning, and synthesis. For explanation generation, we selected a stratified subset of 38 held-out cases spanning low-, intermediate-, and high-risk predictions and ran both generators on the same cases using the same local \texttt{ollama}-served \texttt{llama3.2:3b} model. Both versioned generator scripts wrote to \texttt{results/scirep\_v1/} so reruns would not overwrite the original retained artifacts.

\subsection{Leakage audit and explanation scoring}

The scirep\_v1 audit searched explanation text for explicit outcome or survival language. Any explanation containing terms such as ``actual outcome'', ``survived'', ``survival'', or ``expired'' was flagged and excluded from valid explanation-quality scoring. Retained explanations were evaluated on SHAP alignment (Jaccard overlap with top-3 SHAP features), clinical plausibility, direction consistency, value specificity, guideline grounding, and reasoning depth. Head-to-head comparison between standalone and agentic outputs was restricted to the overlap between the generated subset and the retained SHAP-reviewed patient subset. Confidence intervals for mean quality metrics were estimated by non-parametric bootstrap over the retained overlapping cases.

\section*{Data availability}

The data supporting the findings of this study are available from the corresponding author upon reasonable request. The study also uses the publicly available eICU Collaborative Research Database Demo \cite{pollard2018}.

\section*{Code availability}

The code used for the analyses in this study is available from the corresponding author upon reasonable request.

\section*{Author contributions}

Di Zhu and Chen Xie contributed equally to the review and analysis of the literature and to revision of the manuscript. Ziwei Wang served as the corresponding author. Chen Xie prepared the revised manuscript files. All authors approved the final version of this draft.

\section*{Competing interests}

The authors declare no competing interests.

\section*{Acknowledgements}

This research used the eICU Collaborative Research Database Demo, made available by Philips Healthcare and the MIT Laboratory for Computational Physiology.

\input{main_scirep_v1_refs.tex}
\end{document}

%% file: results/scirep_v1/tables/table1_demographics_scirep_v1.tex
\begin{table}[t]
\centering
\caption{Baseline characteristics of the study cohort}
\label{tab:demographics}
\begin{tabular}{lccc}
\toprule
Variable & Overall (N=2353) & Survivors (N=2162) & Non-Survivors (N=191) \\
\midrule
Age (years) & 63.3 ± 17.7 & 62.5 ± 17.8 & 73.2 ± 12.8 \\
Male sex & 1415 (60.1\%) & 1305 (60.4\%) & 110 (57.6\%) \\
BMI (kg/m²) & 28.6 ± 8.1 & 28.7 ± 8.1 & 27.9 ± 7.7 \\
Heart rate (bpm) & 84.1 ± 16.8 & 83.4 ± 16.4 & 92.4 ± 19.3 \\
MAP (mmHg) & 80.4 ± 18.9 & 81.0 ± 19.1 & 75.8 ± 17.5 \\
SpO2 (\%) & 96.4 ± 3.2 & 96.5 ± 2.9 & 95.2 ± 5.3 \\
Temperature (°C) & 38.6 ± 9.9 & 38.5 ± 9.6 & 39.1 ± 11.4 \\
Respiratory rate (/min) & 19.5 ± 5.0 & 19.2 ± 4.9 & 22.2 ± 5.8 \\
Creatinine (mg/dL) & 1.5 ± 1.5 & 1.4 ± 1.5 & 1.7 ± 1.3 \\
BUN (mg/dL) & 25.0 ± 18.7 & 24.1 ± 18.3 & 35.1 ± 20.3 \\
Lactate (mmol/L) & 2.4 ± 2.7 & 2.0 ± 1.8 & 4.9 ± 4.6 \\
WBC (K/µL) & 11.7 ± 12.5 & 11.2 ± 6.3 & 16.8 ± 38.4 \\
Hemoglobin (g/dL) & 11.1 ± 2.2 & 11.2 ± 2.2 & 10.5 ± 2.3 \\
Platelets (K/µL) & 203.4 ± 93.0 & 204.4 ± 92.5 & 191.5 ± 99.0 \\
Bilirubin (mg/dL) & 1.2 ± 2.4 & 1.2 ± 2.3 & 1.7 ± 3.6 \\
Glucose (mg/dL) & 142.0 ± 64.1 & 141.3 ± 63.8 & 149.4 ± 67.4 \\
ICU LOS (hours) & 62.0 ± 84.5 & 59.2 ± 76.7 & 93.6 ± 142.5 \\
\bottomrule
\end{tabular}
\end{table}

%% file: results/scirep_v1/tables/table2_model_performance_scirep_v1.tex
\begin{table}[t]
\centering
\caption{Bootstrapped discrimination metrics on the held-out test set.}
\label{tab:model_performance}
\begin{tabular}{lcccc}
\toprule
Model & AUROC & AUROC 95\% CI & AUPRC & AUPRC 95\% CI \\
\midrule
Logistic regression & 0.823 & 0.752-0.886 & 0.345 & 0.218-0.506 \\
XGBoost & 0.855 & 0.796-0.906 & 0.332 & 0.217-0.494 \\
\bottomrule
\end{tabular}
\end{table}

%% file: results/scirep_v1/tables/table3_explanation_quality_scirep_v1.tex
\begin{table}[t]
\centering
\caption{Outcome-leakage audit and retained standalone explanation quality metrics.}
\label{tab:llm_quality}
\begin{tabular}{lc}
\toprule
Metric & Value \\
\midrule
LLM source explanations & 38 \\
Flagged for explicit outcome leakage & 1 \\
Retained after audit & 37 \\
SHAP alignment mean (95\% CI) & 0.171 (0.075-0.279) \\
Plausibility mean (95\% CI) & 0.671 (0.571-0.764) \\
Direction consistency & 92.9\% (78.6-100.0\%) \\
Value specificity mean (95\% CI) & 0.143 (0.029-0.300) \\
Guideline grounding mean (95\% CI) & 0.143 (0.071-0.238) \\
\bottomrule
\end{tabular}
\end{table}

%% file: results/scirep_v1/tables/table4_explanation_comparison_scirep_v1.tex
\begin{table}[t]
\centering
\caption{Audited explanation-quality comparison between the standalone LLM and the pre-specified agentic pipeline.}
\label{tab:explanation_comparison}
\begin{tabular}{lcc}
\toprule
Metric & Standalone LLM & Agent pipeline \\
\midrule
Source explanations & 38 & 38 \\
Flagged for leakage & 1 & 0 \\
Retained after audit & 37 & 38 \\
SHAP alignment mean & 0.171 (0.075-0.279) & 0.077 (0.018-0.143) \\
Plausibility mean & 0.671 (0.571-0.764) & 0.700 (0.557-0.843) \\
Direction consistency & 92.9\% (78.6-100.0\%) & 78.6\% (57.1-100.0\%) \\
Value specificity mean & 0.143 (0.029-0.300) & 0.236 (0.093-0.407) \\
Guideline grounding mean & 0.143 (0.071-0.238) & 0.762 (0.571-0.929) \\
\bottomrule
\end{tabular}
\end{table}